\documentclass[letterpaper]{article} 
\usepackage[]{aaai2026}  
\usepackage{times}  
\usepackage{helvet}  
\usepackage{courier}  
\usepackage[hyphens]{url}  
\usepackage{graphicx} 
\usepackage{natbib}  
\usepackage{caption} 
\usepackage{algorithm}
\usepackage{algorithmic}
\usepackage{float}
\usepackage{subfigure}
\usepackage{multirow}
\usepackage{booktabs}
\usepackage{amsmath}
\usepackage{amssymb}
\usepackage{subcaption}
\usepackage[table]{xcolor}
\usepackage{tabularx}
\usepackage{bm}

\usepackage{color}

\usepackage{docmute}
\usepackage{newfloat}
\usepackage{listings}
\DeclareCaptionStyle{ruled}{labelfont=normalfont,labelsep=colon,strut=off} 
\floatstyle{ruled}
\newfloat{listing}{tb}{lst}{}
\floatname{listing}{Listing}
\title{When Eyes and Ears Disagree: Can MLLMs Discern Audio-Visual Confusion?}

\author{
    Qilang Ye\textsuperscript{\rm 1} \textsuperscript{\rm 2}, 
    Wei Zeng\textsuperscript{\rm 2}, 
    Meng Liu\textsuperscript{\rm 2} \textsuperscript{\rm 3}\thanks{Corresponding Authors},
    Jie Zhang \textsuperscript{\rm 4}, 
    Yupeng Hu \textsuperscript{\rm 6}, 
    Zitong Yu \textsuperscript{\rm 4} \textsuperscript{\rm 5}\footnotemark[\value{footnote}], 
    Yu Zhou\textsuperscript{\rm 1} \textsuperscript{\rm 2}\\
}
\affiliations{
    \textsuperscript{\rm 1}VCIP \& TMCC \& DISSec, College of Computer Science \& College of Cryptology and Cyber Science, Nankai University\\
    \textsuperscript{\rm 2}Beijing Zhongguancun Academy\\
    \textsuperscript{\rm 3}Shandong Jianzhu University\\
    \textsuperscript{\rm 4}Great Bay University\\
    \textsuperscript{\rm 5}Dongguan Key Laboratory for Intelligence and Information Technology\\
    \textsuperscript{\rm 6}Shandong University\\
    \{yql, s-zw24\}@bjzgca.edu.cn, mengliu.sdu@gmail.com, jz@stu.cqut.edu.cn, huyupeng@sdu.edu.cn, yuzitong@gbu.edu.cn, yzhou@nankai.edu.cn

}

\usepackage{bibentry}

\begin{document}

\maketitle

\begin{abstract}
\textit{\textbf{Can Multimodal Large Language Models (MLLMs) discern confused objects that are visually present but audio-absent?}} To study this, we introduce a new benchmark, AV-ConfuseBench, which simulates an ``Audio-Visual Confusion'' scene by modifying the corresponding sound of an object in the video, e.g., mute the sounding object and ask MLLMs ``\textit{Is there a/an \{muted-object\} sound}''. Experimental results reveal that MLLMs, such as Qwen2.5-Omni and Gemini 2.5, struggle to discriminate non-existent audio due to visually dominated reasoning. Motivated by this observation, we introduce \textbf{RL-CoMM}, a \textbf{R}einforcement \textbf{L}earning-based \textbf{Co}llaborative \textbf{M}ulti-\textbf{M}LLM that is built upon the Qwen2.5-Omni foundation. RL-CoMM includes two stages: 1) To alleviate visually dominated ambiguities, we introduce an external model, a Large Audio Language Model (LALM), as the reference model to generate audio-only reasoning. Then, we design a Step-wise Reasoning Reward function that enables MLLMs to self-improve audio-visual reasoning with the audio-only reference. 2) To ensure an accurate answer prediction, we introduce Answer-centered Confidence Optimization to reduce the uncertainty of potential heterogeneous reasoning differences. Extensive experiments on audio-visual question answering and audio-visual hallucination show that RL-CoMM improves the accuracy by 10$\sim$30\% over the baseline model with limited training data. Follow: https://github.com/rikeilong/AVConfusion.
\end{abstract}
\section{Introduction}\label{intro}

We perceive things by gathering visual information from our eyes while constantly acquiring knowledge through our hearing. Recent advancements in Multimodal Large Language Models (MLLMs\footnote{In this paper, MLLMs refer to models that can process inputs containing both video and audio.}) \cite{ye2024CAT,videollama2,videollama} show a remarkable ability to understand real-world human language and generate continuum contexts. Furthermore, the wide range of complex visual and audio signals has facilitated the development of MLLMs such as Qwen2.5-Omni \cite{qwenomni} and Gemini 2.5 Pro \cite{gemini}. After fine-tuning with an extensive synchronized video and sound corpus, MLLMs develop a cognitive capacity for audio-visual understanding in areas including automatic speech recognition \cite{hpl1}, audio-visual captioning \cite{anygpt}, and general audio-visual processing \cite{onellm}. Despite the excellent generative capabilities of MLLMs for generic scenarios, studies on Audio-Visual Hallucinations (AVH) \cite{hall2} have revealed that MLLMs are unable to distinguish between the volume and pitch of two pieces of audio. Furthermore, MLLMs are susceptible to hearing fictitious sounds or perceiving imaginary visual objects in cross-modal understanding scenarios \cite{avhbench}. 

\begin{figure}[t]

  \centering
  \includegraphics[scale=0.6]{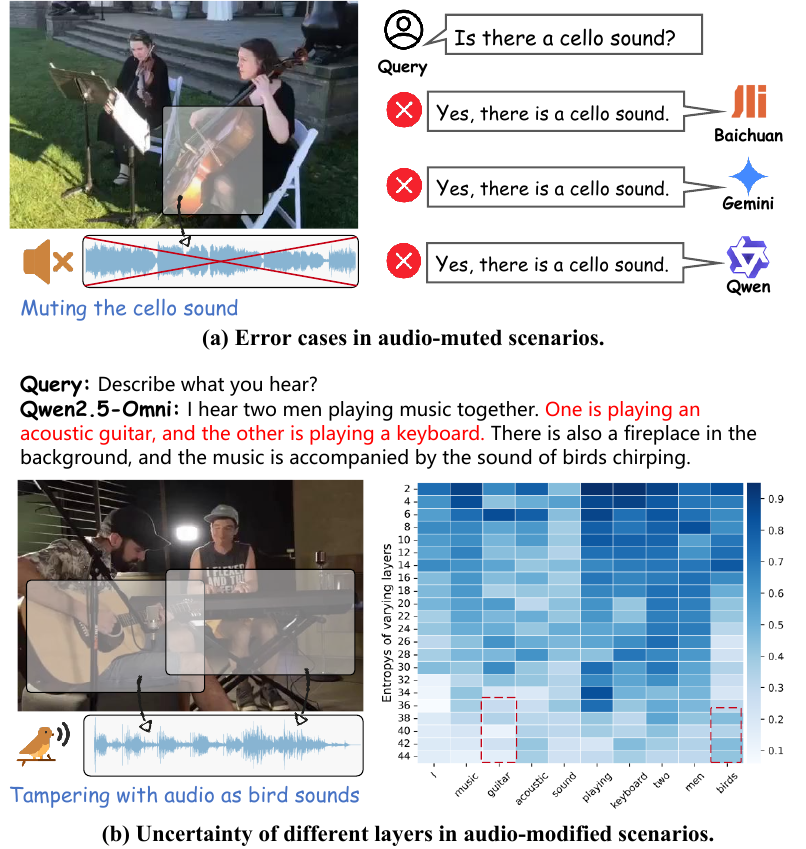}
  \caption{Examples of MLLMs confronting audio-visual confusion.
  }
   \label{fig1}
\end{figure}


In this paper, we further explore: if the given audio-visual information is asymmetric, $\diamondsuit$ \textbf{\textit{Can MLLMs discern confused objects that are visually present but audio-absent?}} We call this phenomenon ``\textbf{Audio-Visual Confusion}'', which refers to the given audio-visual information being inherently asymmetric. As illustrated in Fig. \ref{fig1}, we primarily focus on two settings: \textbf{(a) Test whether MLLMs can recognize that a certain muted object is not making a sound}. In normal cognition, vision and audio are bound, i.e., the presence of some object necessitates the presence of sound feedback. But what if given a damaged video that is visually intact but acoustically crippled, would MLLMs recognize it? For example, we input a ``cello-muted'' video and ask Baichuan-Omni-7B \cite{baichuan}, Gemini 2.5 Pro \cite{gemini}, and Qwen2.5-Omni-7B \cite{qwenomni} if there is a cello sound. All models fail to notice the missing audio but blindly believe in the visual information. Further, we explore \textbf{(b) Test whether MLLMs can balance audio-visual information for reasoning}. For example, we input an extremely unbalanced video (tampering with the music as bird sounds) to Qwen2.5-Omni. Then, we follow an entropy-based metric \cite{entropy-metric,zou2024look} to quantify the uncertainty of the next token in each layer. As shown on the right of Fig. \ref{fig1}, although Qwen2.5-Omni is able to distinguish \textit{bird} sounds, its uncertainty scores remain relatively high in the latter layers, which means the model is still being biased towards hearing \textit{acoustic guitar} and \textit{keyboard} sounds when generating responses.

To further investigate this phenomenon, we propose a mini-benchmark, \textbf{AV-ConfuseBench}, to assess the ability of MLLMs to distinguish audio-visual confusions. It consists of two settings: audio-muted (i.e., muting one of the multiple instruments all the time) and audio-modified (i.e., tampering with the soundtrack of the entire video to be completely out of sync with the theme). Audio-muted setting is tested by giving a query: ``Is there a/an \{muted-object\} sound?'' and models reply yes/no; audio-modified setting aims at evaluating the consistency of the generated content with the audio-visual ground truth.

As shown in Table \ref{avconfusebench}, we find that: 1) None of the MLLMs, especially open-source models, yield the expected results in the audio-muted scenario, which means they struggle to discriminate non-existent audio tokens due to visual influences. 2) While Gemini 2.5 Pro with thinking mode achieves higher performance over the baseline, it remains difficult to think outside of the inherent internal knowledge in multimodal tasks containing video and audio (About 38.36\% of the responses remain affected by visually guided thinking and output errors). 3) MLLMs appear to be insensitive to audio content, and the reasoning is dominated by the presented visual in the audio-modified scenario. This means the generated contexts mostly elaborate visual information.

Motivated by this observation, we introduce \textbf{RL-CoMM}, a \textbf{R}einforcement \textbf{L}earning-based \textbf{Co}llaborative \textbf{M}ulti-\textbf{M}LLMs system that is built upon the Qwen2.5-Omni-3B foundation. RL-CoMM consists of two different stages: 1) Step-wise Reasoning Reward function (Step-RR) and 2) Answer-centered Confidence Optimization (Ans-CO). For the first stage, RL-CoMM introduces an external model, i.e., Large Audio Language Models (LALMs), as the reference model to supplement audio evidence. Then, we design Step-RR based on Group Relative Policy Optimization (GRPO), which uses the extra audio knowledge to incentivize the policy model for audio context reasoning and audio-visual correlation reasoning. For the second stage, Ans-CO can reduces the policy model's uncertainty over its answer predictions to resolve uncertainties arising from potentially heterogeneous reasoning differences. Through extensive experiments on Audio-visual Question Answering (AVQA) tasks and AVH benchmarks, our proposed RL-CoMM improves accuracy by 10$\sim$30\% over the base-LLM only with around 20\% of total training samples. 

\begin{table}[t]
  \centering
\setlength{\tabcolsep}{0.1mm}{
  \begin{tabular}{lcccc}
    \toprule
    \multirow{3}{*}{Model} & \multicolumn{2}{c}{\textbf{Audio-muted}}  & \multicolumn{2}{c}{\textbf{Audio-modified}}\\
    \cmidrule(lr){2-3}\cmidrule(lr){4-5}
    & Acc. ($\uparrow$) & Yes (\%) & A-Acc. ($\uparrow$) & V-Acc. ($\uparrow$) \\
   \rowcolor{gray!30}\multicolumn{5}{c}{\textit{\small{Open-source omni-models}}} \\
   Video-LLaMA2-7B & 2.73 & 97.27 & 0.88 & 3.88\\
   Baichuan-Omni-7B & 5.47 & 94.53 & 1.12 & 4.07\\
   Qwen2.5-Omni-7B & 9.59 & 90.41 & 1.02 & 4.32\\ 
   \midrule
      \rowcolor{gray!30}\multicolumn{5}{c}{\textit{\small{Close-source omni-models}}} \\
   Gemini 2.5 Flash & \underline{28.76}&\underline{71.24}&2.24&\textbf{4.78}\\ 
   Gemini 2.5 Pro & \textbf{68.50} & \textbf{31.50} & \textbf{2.83} & \underline{4.66}\\ 
    \midrule

   Random Choice & 50.00 & 50.00 & - & - \\
  \bottomrule
  \end{tabular}}
  \caption{Results of different MLLMs on AV-ConfuseBench. We evaluate various open/closed-source in two settings, where ``Yes'' is the proportion of answer yes among total responses, ``A-Acc.'', ``V-Acc.'' refer to scoring audio and visual response accuracy on a 0-5 scale using GPT-4.}\label{avconfusebench}
\end{table}

\section{Related Work}

\subsubsection{Large Audio-Visual Language Models.} Drawing inspiration from the remarkable ability of Large Language Models (LLMs) \cite{qwen2, qwen2.5, llama2} to generate coherent language, studies have extended LLMs to other multimodal tasks, e.g., Audio-visual Question Answering (AVQA) \cite{pianoavqa}, and human-centric understanding tasks \cite{ye2024iet, ye2024pose}. Primarily, visual LLMs \cite{ye2025cat+,fusionmamba,liu2018attentive,liu2018cross,lin2025reliable} emphasize the design of elegant bridging methods. In this work, we focus on Omni-LLMs \cite{ye2024CAT,qwenomni,videollama2}, it refers to models that understand both video and audio inputs, which possess multisensory properties that mimic human perception. However, we find that these models trained from synchronized audio-visual data are highly susceptible to complementary modalities.

\subsubsection{Audio-visual Defects in MLLMs.} Hallucination \cite{pope,hall,hall2,hall3,hall4,shu2025semantics} refers to the generation of imaginative textual responses by the model that do not correspond to the input signal. Studies \cite{avhbench} have revealed that such phenomenon is caused by favoring the internal knowledge of the LLM and disregarding the input signal. In this paper, we present another shortage of MLLMs that is similar to hallucinations: ``Audio-Visual Confusion''. Specifically, we test the ability of MLLMs to cope with asymmetric audio-visual information.

\begin{figure*}[]
  \centering
  \includegraphics[scale=0.55]{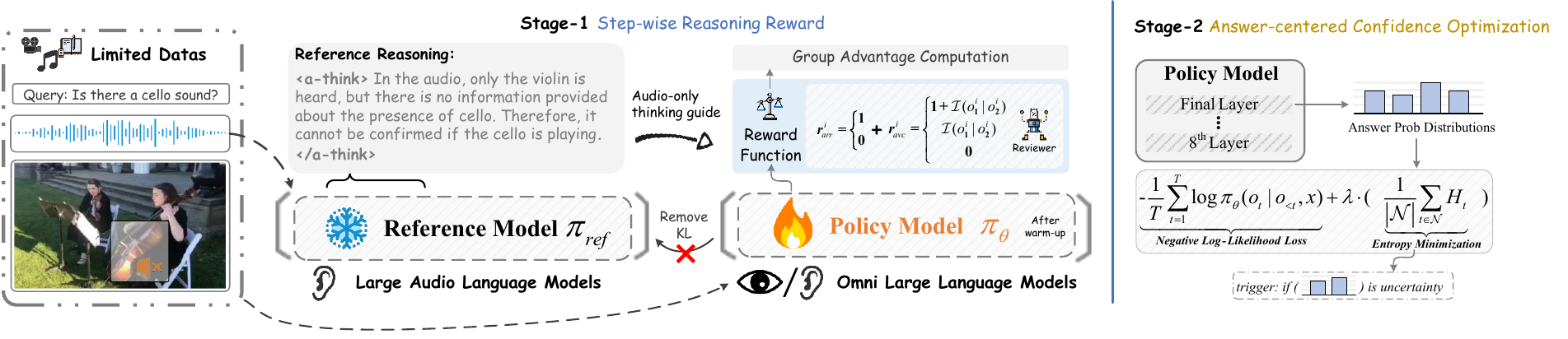}
  \caption{Framework of RL-CoMM, where LALMs serve as the reference model and Omni-LLMs serve as the policy model. Given audio-visual inputs, we first let the LALM generate the reference reasoning for the audio. The policy model is verified by the reviewer (Qwen3 Embedding) to compute group advantages via the Step-wise Reasoning Reward function. Notably, we remove the KL penalty during the policy gradient optimization due to heterogeneous model structure differences. Furthermore, we introduce an Answer-centered Confidence Optimization to reduce uncertainty in the predicted answer of the policy model.
  }
   \label{fig2}
\end{figure*}

\subsubsection{MLLMs with Reinforcement Learning.} Reinforcement Learning (RL) from human feedback \cite{rhlfff,rhlf2}, as an early optimization of language models requires significant human annotation and computational resources. Then, on-policy optimization methods such as DPO \cite{dpo}, and PPO \cite{ppo}, which reward fine-tuned models via computing advantages have achieved excellent outcomes. GRPO \cite{grpo}, as the core optimization algorithm of Deepseek-R1, has advanced the intermediate thinking trajectories in response via KL-penalty and reward model. Recent studies \cite{Liu2025SuperRLRL} have focused on unifying Supervised Fine-Tuning (SFT) \cite{sft} and RL to effectively improve the sensitivity of LLMs to the final output. This paradigm utilizes high quality offline data and online-optimization to interleave the training of models. Unlike prior methods, we improve GRPO by including a heterogeneous reference model, audio-LLMs \cite{qwenaudio}, to complement Omni-LLMs \cite{qwenomni} thinking knowledge. Such a paradigm significantly improves the performance of base-LLMs on AVQA with limited data.

\section{AV-ConfuseBench}
We believe that analyzing the reliability of existing MLLMs under audio-visual confusion can develop more robust models. We provide detailed descriptions and construction of two different settings in AV-ConfuseBench below:

\noindent \textbf{Audio-muted Confusion.} This task is set on masking a particular sound source in a scene where multiple instruments are performing, and it assesses whether visual objects affect the audio understanding of MLLMs. All questions are present in the form of: ``\textit{This is a video of audio corruption where some instrument sound is muted. \textbf{question:} Is there a/an \{muted-object\} sound?}'' and the ground truth is ``\textit{No}''. We mute the sound sources of the collected \textbf{39} videos and yield a total of \textbf{73} Q\&A pairs. The evaluation metrics are accuracy and model response ``\textit{Yes}'' coverage.

\noindent \textbf{Audio-modified Confusion.} This task is designed to modify background sounds to assess whether audio-generated false information affects the discriminative ability of MLLMs. All questions are present in the form of ``\textit{Describe what you see and what you hear}''. We collected \textbf{5} different environmental sounds, including sounds of wind, bird, rain, electric drill, and thunder sounds, to tamper with the background sound of \textbf{20} videos and yield a total of \textbf{100} Q\&A pairs. To ensure the quality of the assessment, all ground truths are manually labeled. The evaluation metrics are AI-assisted \cite{deepseekr1} assessment of the accuracy of generated visual and audio contents to ground truths.

\begin{table}[]
\begin{tabular}{l}
\toprule
\parbox[c]{7.8cm}{\textbf{Prompt for $\bm{\pi_{ref}}$:} You are an assistant to help with hearing. Here is the answer ``\{\textit{ground truth}\}'' to the question ``\{\textit{question}\}''. And your task is: Reasoning what you hear that is useful for answering the question in \textless a-think\textgreater  \textless /a-think\textgreater. The output format should be as follows: \textless a-think\textgreater...\textless /a-think\textgreater} \\
\midrule
\parbox[c]{7.8cm}{\textbf{Prompt for $\bm{\pi_{\theta}}$:} You should see clearly the given video and listen clearly to the given audio. Your tasks are: (1) Reasoning what you hear regarding the question \{\textit{question}\} in \textless a-think\textgreater \textless /a-think\textgreater; (2) Reasoning what you see regarding the question in \textless v-think\textgreater \textless /v-think\textgreater; (3) Output the correct answer in \textless answer\textgreater \textless /answer\textgreater; The output format should be as follows: \textless a-think\textgreater...\textless /a-think\textgreater\textless v-think\textgreater...\textless /v-think\textgreater\textless answer\textgreater...\textless /answer\textgreater }\\
\bottomrule
\end{tabular}
\caption{Prompts for different models. The \textit{reference output} is only present during training.}\label{tab-prompts}
\end{table}

\section{Methodology}

\subsection{Preliminary: RL with Verifiable Rewards}
The verifiable reward function, e.g., GRPO \cite{grpo}, is a direct optimization strategy designed to enhance the reasoning capability of the policy model $\pi_{\theta}(\cdot|\cdot)$. It simplifies the post-training cost by removing the value model, while generating multiple responses $\{o\}$ for the input prompt $q$ to measure correctness. Specifically, the GRPO objective is defined as:
\begin{equation}
\begin{aligned}
\mathcal{L}_{GRPO} &= \mathbb{E}_{[(q,o)\sim D]}[R(\theta)-\beta \text{KL}[\pi_{\theta}(o|q)\parallel \pi_{ref}(o|q)],
\end{aligned}
\end{equation}
where $\pi_{ref}$ is the reference model, $R(\theta)$ denotes the reward function: $\text{min}[\frac{\pi_{\theta}(o|q)}{\pi_{ref}(o|q)}A^i,\text{clip}(\frac{\pi_{\theta}(o|q)}{\pi_{ref}(o|q)};1-1-\epsilon,1+\epsilon)A^i)$, $A^i$ denotes the advantage of the $i$-th response, and the policy model is optimized via updating the parameter $\theta$ with gradient ascent. Notably, we have removed the KL loss during the training period.

\subsection{RL-CoMM}
The framework and process of RL-CoMM are shown in Fig. \ref{fig2} and Algorithm \ref{alg:Framwork}, respectively. We treat the LALM as the reference model $\pi_{ref}$ and the Omni-LLM as the policy model $\pi_{\theta}$, respectively. RL-CoMM consists of two optimization stages. After warm-up, Step-RR is used to optimize reasoning, Ans-CO is used to optimize answers. The multi-modal input data consists of video, audio, and two prompts in the specified format.

\subsubsection{Warm-up for Policy Model.}
To ensure a stable reasoning of the foundation model, we introduce supervised fine-tuning of a small number of datasets before online RL optimization. During the warm-up phase, the policy model is trained on a given multimodal dataset, which consists of questions, videos, audios, and long responses constructed with audio-visual context. This phase primarily specifies that the output content of the policy model can include step-by-step visual reasoning and audio reasoning.

\begin{algorithm}[t]
\caption{The process of RL-CoMM.}
\label{alg:Framwork}
\begin{algorithmic}
\renewcommand{\thealgorithm}{}
\REQUIRE Policy model $\pi_\theta$ after warm-up; Reference model $\pi_{ref}$
\ENSURE Optimized policy model $\pi_\theta$\\
\FOR{each $i \in [1,N]$}
\STATE \textbf{if} \textcolor{gray}{Stage == Step-RR}
    \STATE Generate audio reference reasoning paths {$o_{ref}$}$\sim \pi_{ref}$\
    \STATE Generate $G$ audio-visual reasoning paths {$o$}$^i$$\sim \pi_{\theta}$\
    \STATE Compute rewards {$r_{format}^i$}, {$r_{arr}^i$}, {$r_{avc}^i$} via Eqs. \ref{format_r}-\ref{avc_r}\
    \STATE Optimize policy model $\pi_\theta$ with group advantages {$A$}$^i$\
    \STATE \textbf{else if} \textcolor{gray}{Stage == Ans-CO}
\STATE Clip the answer token {$o_{<t}$}$^i$ in {$o$}$^i$\
\STATE Optimize policy model $\pi_\theta$ wih Ans-CO via Eqs. \ref{Ans-CO}
\ENDFOR
\RETURN Policy model $\pi_\theta$
\end{algorithmic}
\end{algorithm}

\subsubsection{Data Preparation.}
Following the recent GRPO-style training paradigm \cite{omni-r1}, we define two multi-modal training data formats. Table \ref{tab-prompts} shows the prompts input to the reference model and the policy model, respectively. We define three tags \textless a-think\textgreater\textless v-think\textgreater\textless answer\textgreater, which serve to sample the audio thinking content, visual thinking content, and predicted answers of the policy models. Notably, we let the reference model generate question-oriented thinking content based on the ground truth. In this way, we avoid the audio-visual correlation thinking inherent to Omni-LLMs and instead reason the question from a specific perspectives. To restrict the output format of the policy model, we use the format reward during the training process as follows:
\begin{equation}\label{format_r}
{r_{format}^{i}} = \left\{ {\begin{array}{*{20}{l}}
\text{1,{\rm{ if format is correct;}}}\\
\text{0,{\rm{ otherwise}}}
\end{array}} \right.
\end{equation}


\subsubsection{Step-wise Reasoning Reward Function.}
A crucial step in RL is to design effective reward models, also known as reward functions. It updates the policy model at each step through generalized advantage while aligning the preference algorithms to prevent model changes from differing too much from the reference model. Traditional RL reward designs for Omni-LLMs primarily focus on answer accuracy and format consistency \cite{echolnk-r1}, lacking incentives for multimodal reasoning content.

For the multiple-choice AVQA tasks, the model should first deduce the audio-visual content, then choose the correct option. We design Step-RR mainly for correcting the potential visual bias and blending audio-visual perception in the reasoning content of Omni-LLMs. Step-RR includes two types of rule-based rewards, i.e., Audio Reasoning Rationality reward (ARR) $r_{arr}$ and Audio-Visual Correlation (AVC) reward $r_{avc}$. Furthermore, to ensure efficient and accurate reward allocation, we use an off-the-shelf lightweight and powerful text embedding model, i.e., Qwen3 Embedding-0.6B \cite{qwen3embed}, for semantic alignment. Specifically, we define the audio reasoning path to the question $q$ from the reference model $\pi_{ref}$ as $o_{ref}$, the policy model $\pi_{\theta}$ generates a group of $G$ reasoning trajectories for each question $q$ in one rollout $\{o\}^i = (o_{1}^{i},o_{2}^{i},o_{3}^{i})$ containing the audio thinking content, visual thinking content, and predicted answers.

\noindent \textbf{(1) ARR} is used to assess the validity of the policy model for auditory perception $o_{1}^{i}$ by including extra audio knowledge $o_{ref}$ that contributes to the solution. ARR is defined as:
\begin{equation}\label{arr_r}
{r_{arr}^{i}}=\left\{
\begin{array}{lcl}
1  & & {\text{if }\mathcal{S}(o_{1}^{i}|o_{ref})>\omega,o_{3}^{i} = y}\\
0 & & {\text{otherwise}},
\end{array} \right.
\end{equation}
where $\mathcal{S}(o_{1}^{i}|o_{ref})$ denotes function that uses Qwen3 Embedding-0.6B to retrieve semantic similarity score between $o_{1}^{i}$ and $o_{ref}$. We define $\omega=0.8$ and the task defined for Qwen3 Embedding-0.6B is ``\textit{Judge whether the given query $o_{1}^{i}$ is semantically consistent with the provided content $o_{ref}$}''.

\noindent \textbf{(2) AVC} is used to assess the correlation between auditory perception $o_{1}^{i}$ and visual perception $o_{2}^{i}$ to allow for a logical structure to reason about the answer. AVC is defined as:
\begin{equation}\label{avc_r}
{r_{avc}^{i}}=\left\{
\begin{array}{lcl}
1 + \mathcal{I}(o_{1}^{i}|o_{2}^{i}) & & {\text{if } o_{3}^{i} = y,}\\
\mathcal{I}(o_{1}^{i}|o_{2}^{i}) & & {\text{if }o_{3}^{i} \neq \text{null}, \neq y,}\\
0 & & {\text{otherwise}},
\end{array} \right.
\end{equation}
where $\mathcal{I}(o_{1}^{i}|o_{2}^{i})$ denotes function that uses Qwen3 Embedding-0.6B to retrieve coherent score between $o_{1}^{i}$ and $o_{2}^{i}$. We include a soft-matching mechanism to ensure a rational reasoning process based on predicted answers. The task defined for Qwen3 Embedding-0.6B is ``\textit{Given a query $o_{1}^{i}$, retrieve semantically coherent content $o_{2}^{i}$}''.

\begin{table*}[t]\centering
\setlength{\tabcolsep}{3mm}{
\begin{tabular}{l|cccccc|c}
\toprule
\multirow{2}{*}{Method} &  \multicolumn{6}{c|}{\textbf{Music-AVQA} (Audio-visual)}  & \textbf{AVQA} \\
& Exist & Localis & Count & Comp & Temp & Avg. & Avg. \\
\midrule
\rowcolor{gray!30}\multicolumn{8}{l}{\textit{\small{Specialized models}}} \\
PSTP-Net &76.18&73.23&71.80&71.79&69.00&72.57&90.20 \\
CAD & 83.42&73.97&76.37&74.88&76.16&76.96&92.20 \\
\rowcolor{gray!30}\multicolumn{8}{l}{\textit{\small{LLM-based models}}} \\
Video-LLaMA&62.16&41.86&42.23&44.98&24.76&47.93&71.69\\
One-LLM&68.71&50.83&67.57&50.76&41.02&57.86&88.75\\
\midrule
Qwen2.5-Omni-3B&60.02&53.84&61.29&58.16&46.57&54.95&83.78\\
+ SFT&73.67&74.09&75.43&68.47&60.44&70.41&90.41\\
+ GRPO&77.69&71.10&67.33&64.23&70.14&70.05&85.31\\
\rowcolor{blue!10}\textbf{+ RL-CoMM (ours)}&\textbf{85.61}&\textbf{76.68}&\textbf{84.08}&\textbf{70.74}&\textbf{76.30}&\textbf{79.46}&\textbf{95.87}\\
$\Delta$&($\uparrow$ \textcolor{blue}{25.59})&($\uparrow$ \textcolor{blue}{22.84})&($\uparrow$ \textcolor{blue}{22.79})&($\uparrow$ \textcolor{blue}{12.58})&($\uparrow$ \textcolor{blue}{29.73})&($\uparrow$ \textcolor{blue}{24.51})&($\uparrow$ \textcolor{blue}{12.09})\\
\bottomrule
\end{tabular}}
\caption{Results on the Music-AVQA and AVQA. Exist, Localis, etc. represent the accuracy in the subtasks of this benchmark.}\label{MUSIC-AVQA}
\end{table*}

\subsubsection{Group Advantage Computation.}
The overall reward consists of format rewards $r_{format}$, audio reasoning rationality reward $r_{arr}$, and audio-visual correlation reward $r_{avc}$. Each reasoning path in a generated group can be computed as $r^{i} = r_{format}+r_{arr}+r_{avc}$, i.e., to yield the group advantages $\{r^1,r^2,...,r^G\}$. Then, we follow the normalization formula \cite{grpo} and define as:
\begin{equation}
    A^i=\frac{r^i-\text{mean}(\{r^1,r^2,...,r^G\})}{\text{std}(\{r^1,r^2,...,r^G\})}.
\end{equation}

\subsubsection{Answer-centered Confidence Optimization.}
Inspired by Entropy Minimization \cite{oneshotem}, we incorporate the Answer-centered Confidence Optimization (Ans-CO) to address the uncertainty associated with potentially heterogeneous reasoning differences. Notably, to ensure fitting the labeled data, we introduce Negative Log-Likelihood (NLL) loss to smooth the objective learning. Specifically, we define $\mathcal{N}=\{t|t>T_{prompt+think}\}$ for cropping the answer portion to avoid computation on the prompt and the think, where $T_{prompt+think}$ denotes the token length of the prompt and generated think content. Then we define Ans-CO as:
\begin{equation}\label{Ans-CO}
    \mathcal{L}_{OP}=\underbrace { - \frac{1}{T}\sum\limits_{t = 1}^T {\log {\pi _\theta }({o_t}|{o_{ < t}},x)} }_{\text{\small Negative Log - Likelihood Loss}} + \lambda  \cdot (\underbrace {\frac{1}{{\left| {\cal N} \right|}}\sum\limits_{t \in {\cal N}} {{H_t}} }_{\text{\small Entropy Minimization}}),
\end{equation}
where ${H_t} =  - \sum\nolimits_{v \in {\cal V}} {{\pi _\theta }(v|{y_{ < t}},x)\log {\pi _\theta }(v|{y_{ < t}},x)}$, $x$ denotes the input, $\mathcal{V}$ is the vocabulary of model $\pi_\theta$. $\lambda$ is the hyperparameter used to avoid overconfidence that leads to decreased generalization. Notably, we follow the entropy-based metric \cite{entropy-metric,zou2024look} to quantify the uncertainty $u$ of the predicted answer, and set $\lambda=0$ when $u > 0.75$. 


\section{Experiments}

\subsection{Implementation Details and Datasets}
RL-CoMM is built upon the Qwen2.5-Omni-3B foundation and consists of three training phases: warm-up via fine-tuning, GRPO-style policy optimization, and Ans-CO. All experiments are conducted on 8 NVIDIA A800 GPUs. Warm-up training is performed on our customized dataset containing 100 high-quality Q\&A pairs with LLaMA-Factory\footnote{https://github.com/hiyouga/LLaMA-Factory/tree/main}. All data used in the warm-up phase strictly adheres to the rules for preventing data leakage. Then, we employ a few-shot learning approach for Step-RR and Ans-CO with limited training samples drawn from the Music-AVQA and AVQA datasets. The hyperparameters $\lambda=0.5$ and $\omega=0.8$ are set by default.

To demonstrate the superiority of RL-CoMM in generic audio-visual scenarios, we conducted experiments on two types of tasks: AVQA (Music-AVQA \cite{musicavqa}, Music-AVQA-R \cite{musicavqar}, and AVQA \cite{avqa}) and AVH (AVHBench \cite{avhbench}).



\begin{table*}[t]\centering
\setlength{\tabcolsep}{2.2mm}{
\begin{tabular}{lccccccccccc}
\toprule
\multirow{2}{*}{Method} & \multicolumn{2}{c}{Exist} & \multicolumn{2}{c}{Localis} & \multicolumn{2}{c}{Count} & \multicolumn{2}{c}{Comp} & \multicolumn{2}{c}{Temp} & \multirow{2}{*}{Avg.} \\
\cmidrule(lr){2-3}\cmidrule(lr){4-5}\cmidrule(lr){6-7}\cmidrule(lr){8-9}\cmidrule(lr){10-11}
& H & T & H & T & H & T & H & T & H & T &\\
\midrule
\rowcolor{gray!30}\multicolumn{12}{l}{\textit{\small{Specialized models}}} \\
LAVisH & 63.17&66.68&30.11&43.80&63.77&26.51&56.31&63.46&50.79&42.85&59.25\\
MCCD & 77.22&67.58&55.15&82.23&70.12&39.83&61.26&58.17&43.67&58.33&66.95 \\
\rowcolor{gray!30}\multicolumn{12}{l}{\textit{\small{LLM-based models}}} \\
Qwen2.5-Omni-3B&61.42&64.58&60.21&68.21&64.91&61.32&49.37&64.76&55.34&68.02&69.43\\
+ GRPO&79.43&80.21&73.21&71.72&71.02&76.94&58.44&70.50&58.49&70.78&71.43\\
\rowcolor{blue!10}\textbf{+ RL-CoMM (ours)}&\textbf{85.98}&\textbf{88.67}&\textbf{81.63}&\textbf{74.28}&\textbf{79.58}&\textbf{83.48}&\textbf{66.89}&\textbf{75.25}&\textbf{63.41}&\textbf{76.08}&\textbf{79.95}\\
\bottomrule
\end{tabular}}
\caption{Results on the Music-AVQA-R test split. H and T denote the head and tail accuracy.}\label{MUSIC-AVQA-R}
\end{table*}

\begin{table*}[!h]\centering
\setlength{\tabcolsep}{0.8mm}{
\begin{tabular}{lccccccccc}
\toprule
\multirow{2}{*}{Method} & \multicolumn{3}{c}{Audio-driven Video Hallucination} & \multicolumn{3}{c}{Video-driven Audio Hallucination} & \multicolumn{3}{c}{Audio-visual Matching} \\
\cmidrule(lr){2-4}\cmidrule{5-7}\cmidrule{8-10}
& Acc. $(\uparrow)$ & Precision $(\uparrow)$&F1 $(\uparrow)$& Acc. $(\uparrow)$ & Precision $(\uparrow)$&F1 $(\uparrow)$& Acc. $(\uparrow)$ & Precision $(\uparrow)$&F1 $(\uparrow)$\\
\midrule
Video-LLaMA&50.1&50.1&66.7&50.2&50.2&66.9&50.0&50.0&66.7\\
ChatBridge&52.9&70.9&48.9&32.8&60.0&39.8&29.9&48.3&33.9\\
PandaGPT&58.5&55.3&68.8&61.3&57.4&69.1&51.2&53.6&27.0\\
OneLLM&53.7&58.6&49.8&44.3&50.2&49.8&\textbf{60.1}&\textbf{67.7}&64.6\\
\midrule
Qwen2.5-Omni-3B&65.85&78.82&79.41&59.65&64.56&74.73&48.77&50.69&65.57\\
+ GRPO&72.98&80.49&81.04&62.84&65.74&75.70&49.73&51.18&66.01\\
\rowcolor{blue!10}\textbf{+ RL-CoMM (ours)}&\textbf{78.96}&\textbf{88.20}&\textbf{88.24}&\textbf{65.63}&\textbf{68.54}&\textbf{79.25}&\underline{51.85}&\underline{53.42}&\textbf{68.42}\\
\bottomrule
\end{tabular}}
\caption{Evaluation results on AVHBench. Acc. denotes the accuracy, and Yes (\%) is the proportion of ``Yes'' answers among total responses. Notably, the tests are performed without introducing the officially provided caption inputs.}\label{avhbench}
\end{table*}

\subsection{Results in AVQA}
\noindent\textbf{Music-AVQA} and \textbf{AVQA} contain 9129 and 57,300 test Q\&A pairs, respectively. Table \ref{MUSIC-AVQA} shows that specialized models such as PSTP-Net \cite{pstp-net} and CAD \cite{CAD} both outperform LLM-based approaches. However, while such models have extremely strong in-domain performance, the shortcoming of a lack of prior knowledge still restricts them from extending to unseen scenarios. For the LLM-based models, we first test the multi-modal comprehension methods, i.e, Video-LLaMA \cite{videollama} and One-LLM \cite{onellm}. Both models are unable to achieve superior results, which proves the necessity of strong audio-visual correlation in the AVQA tasks. On the other hand, Qwen2.5-Omni-3B still falls short of the results achieved by specialized models, despite the overall improvement in performance with SFT. In contrast, RL-CoMM achieves significant improvements in all subtasks compared to the base model. It clearly demonstrates that the proposed reward optimization gives a boost to the reasoning ability of Omni-LLMs.

\noindent\textbf{Music-AVQA-R} is expanded based on Music-AVQA, which includes 211,572 test QA pairs, and RL-CoMM is tested directly without additional training data. The result in Table \ref{MUSIC-AVQA-R} shows that: Specialized models LAVisH \cite{LAVISH}, MCCD \cite{musicavqar} fail to demonstrate strong generalization. In contrast, RL-CoMM unleashes the audio-visual reasoning capabilities of Qwen2.5 Omni and achieves significant improvements in cross-task performance.

\begin{figure}[t]
	\centering
	\begin{minipage}[c]{0.23\textwidth}
		\centering
		\includegraphics[width=\textwidth]{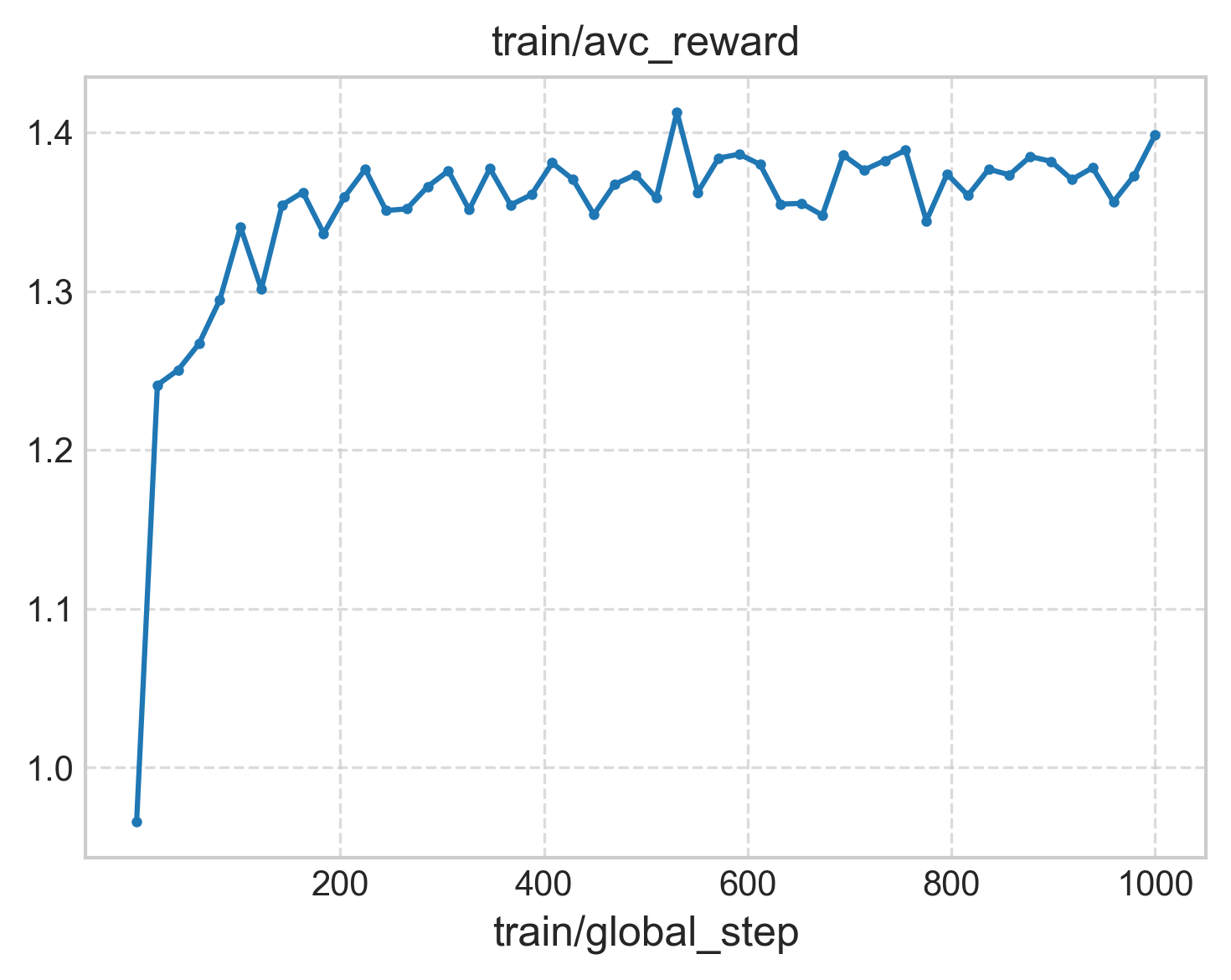}
		\label{fig_E2_3}
	\end{minipage}
    \begin{minipage}[c]{0.23\textwidth}
		\centering
		\includegraphics[width=\textwidth]{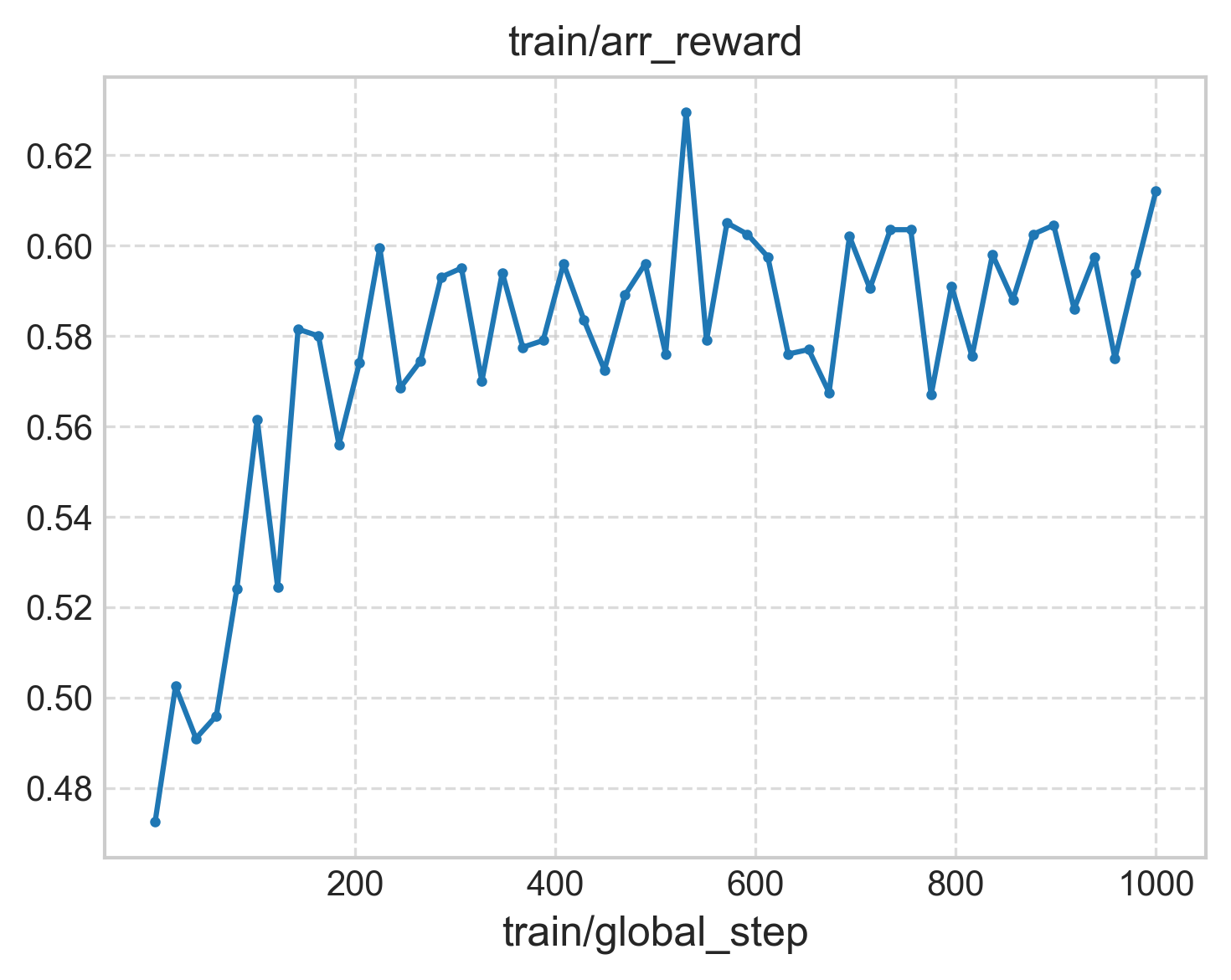}
		\label{fig_E2_2}
	\end{minipage} 
	\caption{Training dynamics of RL-CoMM with Step-wise Reasoning Optimization. The graph below shows the variation of the AVC reward over global steps; the graph above shows the variation of the ARR reward over global steps.}
	\label{fig_dynamic_scores}
\end{figure}

\begin{figure*}
	\centering
		\includegraphics[scale=0.53]{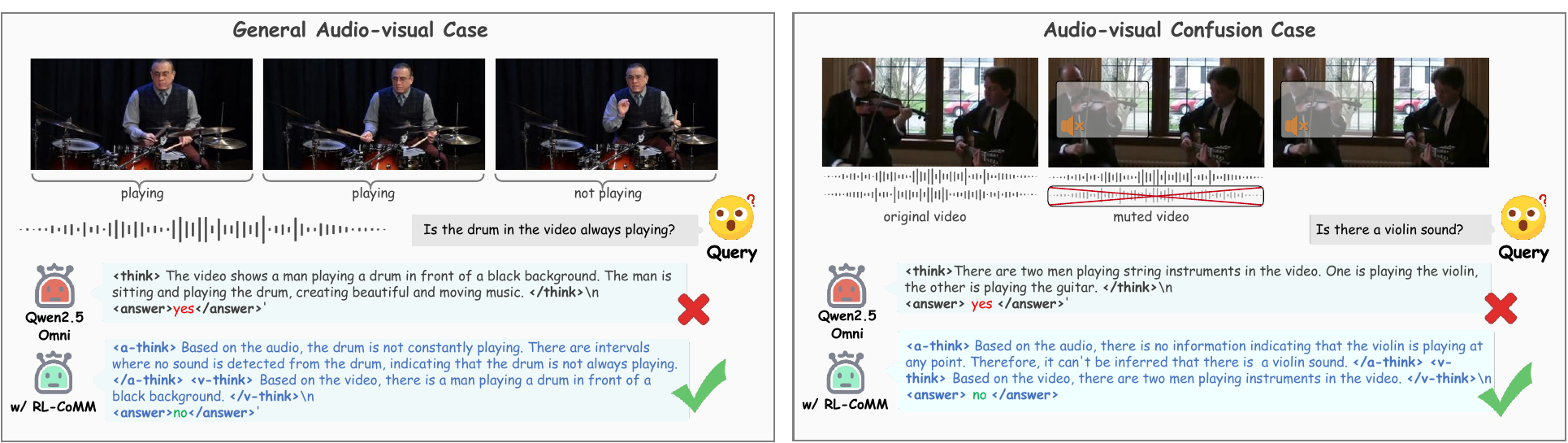}
\caption{Examples of general audio-visual scenes and our designed audio-visual confusion scenes. Questions are answered in the form of yes or no, where the audio information may be intermittent or blocked out.}
	\label{fig:visual}
\end{figure*}

\subsection{Results in AVH}
We mainly explore the improvements that RL-CoMM brings to the base model in three tasks, which are Audio-driven Video Hallucination, Video-driven Audio Hallucination, and Audio-visual Matching tasks. Common open-source models that understand audiovisual content, such as Video-LLaMA \cite{videollama}, ChatBridge \cite{chatbridge}, PandaGPT \cite{pandagpt}, and OneLLM \cite{onellm}, perform poorly and are overconfident in hallucinatory objects. In contrast, Qwen2.5-Omni outperforms the appeal model, whether in audio or visually guided hallucinations, demonstrating strong audio-visual correlation. Moreover, with the reasoning reward optimization, RL-CoMM brings the model up to 13.11\% improvement in accuracy, substantially outperforming models such as OneLLM by 20 to 25\%. Unfortunately, we noticed that performance on the audio-visual matching task failed to optimal. However, online policy optimization indeed reduces the impact of cross-modal hallucinations, motivating us to improve the reward model for audio-visual combinations. 

\begin{table}[t]
  \centering
\setlength{\tabcolsep}{0.1mm}{
  \begin{tabular}{lcccc}
    \toprule
    \multirow{3}{*}{Model} & \multicolumn{2}{c}{\textbf{Audio-muted}}  & \multicolumn{2}{c}{\textbf{Audio-modified}}\\
    \cmidrule(lr){2-3}\cmidrule(lr){4-5}
    & Acc. ($\uparrow$) & Yes (\%) & A-Acc. ($\uparrow$) & V-Acc. ($\uparrow$) \\
   \midrule
   Qwen2.5-Omni-3B & 8.22 & 91.78 & 1.14 & 4.10\\ 
   \midrule
   + SFT & 5.48 & 94.52 & - & - \\
   + GRPO & 15.07 & 84.93 & 1.84 & 4.47 \\
   \textbf{+ RL-CoMM (ours)} & \textbf{27.40} & \textbf{72.60} & \textbf{2.36} & \textbf{4.54} \\
  \bottomrule
  \end{tabular}}
  \caption{Results of different training strategies based on Qwen2.5-Omni-3B in AV-ConfuseBench.}\label{tab-avconfusebench}
\end{table}

\begin{table}[t]\centering
\setlength{\tabcolsep}{1.2mm}{
\begin{tabular}{lc}
\toprule
Method& Avg.\\
\midrule
Qwen2.5-Omni-3B &54.95\\
w/ Format + Accuracy &70.05\\
w/ Format + Step-wise Reasoning &74.49\\
\textbf{w/ Format + Step-wise Reasoning + Ans-CO} &\textbf{79.46}\\
\bottomrule
\end{tabular}}
\caption{Effectiveness of the step-wise reasoning rewards and Ans-CO on the Mean Accuracy of Music-AVQA.}\label{tab_ab1}
\end{table}


\subsection{Analysis in AV-ConfuseBench}
In fact, the core problem that leads to the shortcomings of MLLMs in audio-visual confusion is that the model does not think about the context of the two modalities independently. As shown in Table \ref{tab-avconfusebench}, using SFT to train the model reduces 2.74\%, while the training strategy by stimulating the model to think brings different levels of improvement to the model (6.85\% and 0.7 for GRPO, and 19.18\% and 1.22 for RL-CoMM, respectively). We believe that forcing models to reflect and trial-and-error in training can mimic human thinking to solve difficult audio-visual tasks such as hallucinations and confusions. Regarding the comparison with other methods, while RL-CoMM lags behind the baseline by a relatively large margin, it achieves a twofold improvement over smaller-scale Qwen models, and even approaches the performance of Gemini 2.5 Flash.

\subsection{Training Curves of Reward Optimization}
Exploring the variation of reward scores during online policy optimization can more clearly validate the effectiveness of reward models. As shown in Fig. \ref{fig_dynamic_scores}, the AVC reward shows a smooth upward trend, indicating that the proposed reward model continuously motivates the model to obtain stronger audio-visual associations through groupwise feedback. On the other hand, we observe that the ARR rewards driven by the external model fluctuate considerably in the middle of the training, and the peaks fail to reach the desired results. We argue that Omni-LLMs such as Qwen2.5-Omni, while learning bi-directional audio-visual generation during the warm-up phase, still struggle to make visually unaffected reasoning about audio content. This further proves the need for the stimulus model to independently ``hear'' and ``see''.

\subsection{Ablation Studies}
\subsubsection{Effects of Step-RR and Ans-CO.} We conduct ablation study for Step-RR function, and Ans-CO on Qwen2.5-Omni-3B over the Music-AVQA benchmark. As shown in Table \ref{tab_ab1}, the base GRPO pattern includes rewards and accuracy rewards that provide a slight boost to the model, yet are still not comparable to specialized models. When involving rewards for audio-visual reasoning, it improves the baseline model by 19.54\%, resulting in a 4.44\% increase compared to GRPO. In addition, with the optimization of the confidence in the predicted answers, there is a substantial improvement of 26.51\% over the baseline. We believe that combining two optimization objectives can further enhance the robustness of the model for multimodal understanding.


\subsection{Qualitative Examples}
As shown in Fig. \ref{fig:visual}, we compare the case of Qwen2.5-Omni-3B before and after using RL-CoMM in two audio-visual scenarios. While Qwen2.5 Omni can generate answers through a prescribed format, the reasoning process is visually biased and ignores audio context. Even in the common audio-visual task (shown on the left), the model does not pay attention to the intermittent sound of the drums on the audio but relies directly on the visual and answers ``yes''. In contrast, RL-CoMM can bring a more harmonized audio-visual reasoning to Qwen2.5-Omni. For example, RL-CoMM facilitates model reasoning on audio and visual, respectively, and leans on the audio reasoning based on the question type to answer ``no''. In the audio-visual confusion scene, RL-CoMM simulates the human senses to perceive visual and audio information independently and integrate them, thus being able to avoid false prediction.

\section{Conclusion}

Our primary focus is on ``\textit{Can MLLMs discern confused objects that are visually present but audio-absent?}'' The proposed AV-ConfuseBench reveals that most open-source or closed-source models fail to recognize information asymmetry in both vision and audio. Motivated by this observation, we propose RL-CoMM, which includes two novel optimization strategies. As a result, RL-CoMM significantly improves the performance of Qwen2.5-Omni-3B on both AVQA and AVH, under limited training data.


\section{Acknowledgments}
This research was funded by the Beijing Zhongguancun Academy (Grant No. 20240306), the National Natural Science Foundation of China (Grant No. 62376266 and 62406318), the CCF-Tencent Rhino-Bird Open Research Fund, and the National Natural Science Foundation of China (Grant No. 62576076). 

\bigskip

\bibliography{aaai2026}

@String(CVPR= {IEEE Conf. Comput. Vis. Pattern Recog.})

@String(ECCV= {Eur. Conf. Comput. Vis.})

@String(ACMMM= {ACM Int. Conf. Multimedia})

@String(ICLR = {Int. Conf. Learn. Represent.})

@String(AAAI = {AAAI})

@String(CVPR  = {CVPR})

@String(ECCV  = {ECCV})

@String(ACMMM = {ACM MM})

@String(ICLR  = {ICLR})

@article{shu2025semantics,
  title={When Semantics Mislead Vision: Mitigating Large Multimodal Models Hallucinations in Scene Text Spotting and Understanding},
  author={Shu, Yan and Lin, Hangui and Liu, Yexin and Zhang, Yan and Zeng, Gangyan and Li, Yan and Zhou, Yu and Lim, Ser-Nam and Yang, Harry and Sebe, Nicu},
  journal={arXiv preprint arXiv:2506.05551},
  year={2025}
}

@inproceedings{liu2018attentive,
  title={Attentive moment retrieval in videos},
  author={Liu, Meng and Wang, Xiang and Nie, Liqiang and He, Xiangnan and Chen, Baoquan and Chua, Tat-Seng},
  booktitle={The 41st international ACM SIGIR conference on research \& development in information retrieval},
  pages={15--24},
  year={2018}
}

@inproceedings{liu2018cross,
  title={Cross-modal moment localization in videos},
  author={Liu, Meng and Wang, Xiang and Nie, Liqiang and Tian, Qi and Chen, Baoquan and Chua, Tat-Seng},
  booktitle={Proceedings of the 26th ACM international conference on Multimedia},
  pages={843--851},
  year={2018}
}

@inproceedings{musicavqa,
 author = {Li, Guangyao and Wei, Yake and Tian, Yapeng and Xu, Chenliang and Wen, Ji-Rong and Hu, Di},
 title = {Learning to answer questions in dynamic audio-visual scenarios.},
 booktitle = CVPR,
 pages = {19086–19096},
 year = 2022
}

@inproceedings{avqa,
  author       = {Pinci Yang and
                  Xin Wang and
                  Xuguang Duan and
                  Hong Chen and
                  Runze Hou and
                  Cong Jin and
                  Wenwu Zhu},
  title        = {{AVQA:} {A} Dataset for Audio-Visual Question Answering on Videos},
  booktitle    = ACMMM,
  pages        = {3480--3491},
  publisher    = {{ACM}},
  year         = {2022},
}

@inproceedings{pianoavqa,
 author={Yun, Heeseung and Yu, Yang and Yang, Woosung and Lee, Kang-Il and Kim, Gun-Hee}, 
 title={Pano-AVQA: Grounded Audio-Visual Question Answering on 360$^\circ$ Videos}, 
 booktitle = CVPR,
 pages = {2031–2041},
 year = 2021
}

@inproceedings{videollama,
  author       = {Hang Zhang and
                  Xin Li and
                  Lidong Bing},
  title        = {Video-LLaMA: An Instruction-tuned Audio-Visual Language Model for
                  Video Understanding},
  booktitle    = {Proceedings of the Empirical Methods in Natural
                  Language Processing, {EMNLP}},
  pages        = {543--553},
  year         = {2023},
}

@article{baichuan,
  author       = {Yadong Li and
                  Haoze Sun and
                  Mingan Lin and
                  et al.},
  title        = {Baichuan-Omni Technical Report},
  journal      = {CoRR},
  volume       = {abs/2410.08565},
  year         = {2024},
  eprinttype    = {arXiv},
  eprint       = {2410.08565},
}

@article{onellm,
  author       = {Jiaming Han and
                  Kaixiong Gong and
                  Yiyuan Zhang and
                  Jiaqi Wang and
                  Kaipeng Zhang and
                  Dahua Lin and
                  Yu Qiao and
                  Peng Gao and
                  Xiangyu Yue},
  title        = {OneLLM: One Framework to Align All Modalities with Language},
  journal      = {CoRR},
  volume       = {abs/2312.03700},
  year         = {2023},
  eprinttype    = {arXiv},
  eprint       = {2312.03700},

}

@article{sft,
  author       = {Shengyu Zhang and
                  Linfeng Dong and
                  Xiaoya Li and
                  Sen Zhang and
                  Xiaofei Sun and
                  Shuhe Wang and
                  Jiwei Li and
                  Runyi Hu and
                  Tianwei Zhang and
                  Fei Wu and
                  Guoyin Wang},
  title        = {Instruction Tuning for Large Language Models: {A} Survey},
  journal      = {CoRR},
  volume       = {abs/2308.10792},
  year         = {2023},
  eprinttype    = {arXiv},
  eprint       = {2308.10792},
}

@inproceedings{rhlfff,
  author       = {Long Ouyang and
                  Jeffrey Wu and
                  Xu Jiang and
                  Diogo Almeida and
                  Carroll L. Wainwright and
                  Pamela Mishkin and
                  Chong Zhang and
                  Sandhini Agarwal and
                  Katarina Slama and
                  Alex Ray and
                  John Schulman and
                  Jacob Hilton and
                  Fraser Kelton and
                  Luke Miller and
                  Maddie Simens and
                  Amanda Askell and
                  Peter Welinder and
                  Paul F. Christiano and
                  Jan Leike and
                  Ryan Lowe},
  title        = {Training language models to follow instructions with human feedback},
  booktitle    = {Advances in Neural Information Processing Systems, {NeurIPS}},
  year         = {2022},
}

@article{videollama2,
  author       = {Zesen Cheng and
                  Sicong Leng and
                  Hang Zhang and
                  Yifei Xin and
                  Xin Li and
                  Guanzheng Chen and
                  Yongxin Zhu and
                  Wenqi Zhang and
                  Ziyang Luo and
                  Deli Zhao and
                  Lidong Bing},
  title        = {VideoLLaMA 2: Advancing Spatial-Temporal Modeling and Audio Understanding
                  in Video-LLMs},
  journal      = {CoRR},
  volume       = {abs/2406.07476},
  year         = {2024},
  eprinttype    = {arXiv},
  eprint       = {2406.07476},
}

@article{qwen2,
  author       = {An Yang and
                  Baosong Yang and
                  Binyuan Hui and
                  et al.},
  title        = {Qwen2 Technical Report},
  journal      = {CoRR},
  volume       = {abs/2407.10671},
  year         = {2024},
  eprinttype    = {arXiv},
  eprint       = {2407.10671},
}

@article{pandagpt,
  author       = {Yixuan Su and
                  Tian Lan and
                  Huayang Li and
                  Jialu Xu and
                  Yan Wang and
                  Deng Cai},
  title        = {PandaGPT: One Model To Instruction-Follow Them All},
  journal      = {CoRR},
  volume       = {abs/2305.16355},
  year         = {2023},
  eprinttype    = {arXiv},
  eprint       = {2305.16355},
}

@inproceedings{musicavqar,
  author       = {Jie Ma and
                  Min Hu and
                  Pinghui Wang and
                  Wangchun Sun and
                  Lingyun Song and
                  Hongbin Pei and
                  Jun Liu and
                  Youtian Du},
  title        = {Look, Listen, and Answer: Overcoming Biases for Audio-Visual Question
                  Answering},
  booktitle    = {Advances in Neural Information Processing Systems 38: Annual Conference
                  on Neural Information Processing Systems 2024, NeurIPS 2024, Vancouver,
                  BC, Canada, December 10 - 15, 2024},
  year         = {2024},
}

@article{oneshotem,
  author       = {Zitian Gao and
                  Lynx Chen and
                  Joey Zhou and
                  Bryan Dai},
  title        = {One-shot Entropy Minimization},
  journal      = {CoRR},
  volume       = {abs/2505.20282},
  year         = {2025},
  eprinttype    = {arXiv},
  eprint       = {2505.20282},
}

@article{qwen3embed,
  author       = {Yanzhao Zhang and
                  Mingxin Li and
                  Dingkun Long and
                  et al.},
  title        = {Qwen3 Embedding: Advancing Text Embedding and Reranking Through Foundation
                  Models},
  journal      = {CoRR},
  volume       = {abs/2506.05176},
  year         = {2025},
  eprinttype    = {arXiv},
  eprint       = {2506.05176},
}

@inproceedings{hall3,
  author       = {Anisha Gunjal and
                  Jihan Yin and
                  Erhan Bas},
  editor       = {Michael J. Wooldridge and
                  Jennifer G. Dy and
                  Sriraam Natarajan},
  title        = {Detecting and Preventing Hallucinations in Large Vision Language Models},
  booktitle    = {Thirty-Eighth {AAAI} Conference on Artificial Intelligence, {AAAI}
                  2024, Thirty-Sixth Conference on Innovative Applications of Artificial
                  Intelligence, {IAAI} 2024, Fourteenth Symposium on Educational Advances
                  in Artificial Intelligence, {EAAI} 2014, February 20-27, 2024, Vancouver,
                  Canada},
  pages        = {18135--18143},
  publisher    = {{AAAI} Press},
  year         = {2024},}

@article{hall4,
  author       = {Hanchao Liu and
                  Wenyuan Xue and
                  Yifei Chen and
                  Dapeng Chen and
                  Xiutian Zhao and
                  Ke Wang and
                  Liping Hou and
                  Rongjun Li and
                  Wei Peng},
  title        = {A Survey on Hallucination in Large Vision-Language Models},
  journal      = {CoRR},
  volume       = {abs/2402.00253},
  year         = {2024},
  eprinttype    = {arXiv},
  eprint       = {2402.00253},}

@article{qwenaudio,
  author       = {Yunfei Chu and
                  Jin Xu and
                  Qian Yang and
                  et al.},
  title        = {Qwen2-Audio Technical Report},
  journal      = {CoRR},
  volume       = {abs/2407.10759},
  year         = {2024},
  eprinttype    = {arXiv},
  eprint       = {2407.10759},
}

@article{fusionmamba,
  title={Fusionmamba: Dynamic feature enhancement for multimodal image fusion with mamba},
  author={Xie, Xinyu and Cui, Yawen and Tan, Tao and Zheng, Xubin and Yu, Zitong},
  journal={Visual Intelligence},
  volume={2},
  number={1},
  pages={37},
  year={2024}
}

@article{qwen2.5,
  author       = {An Yang and
                  Baosong Yang and
                  Beichen Zhang and
                  et al.},
  title        = {Qwen2.5 Technical Report},
  journal      = {CoRR},
  volume       = {abs/2412.15115},
  year         = {2024},
  eprinttype    = {arXiv},
  eprint       = {2412.15115},
}

@article{hall2,
  author       = {Kaixiong Gong and
                  Kaituo Feng and
                  Bohao Li and
                  Yibing Wang and
                  Mofan Cheng and
                  Shijia Yang and
                  Jiaming Han and
                  Benyou Wang and
                  Yutong Bai and
                  Zhuoran Yang and
                  Xiangyu Yue},
  title        = {AV-Odyssey Bench: Can Your Multimodal LLMs Really Understand Audio-Visual
                  Information?},
  journal      = {CoRR},
  volume       = {abs/2412.02611},
  year         = {2024},
  eprinttype    = {arXiv},
  eprint       = {2412.02611},
}

@article{chatbridge,
  author       = {Zijia Zhao and
                  Longteng Guo and
                  Tongtian Yue and
                  Sihan Chen and
                  Shuai Shao and
                  Xinxin Zhu and
                  Zehuan Yuan and
                  Jing Liu},
  title        = {ChatBridge: Bridging Modalities with Large Language Model as a Language
                  Catalyst},
  journal      = {CoRR},
  volume       = {abs/2305.16103},
  year         = {2023},
  eprinttype    = {arXiv},
  eprint       = {2305.16103},
}

@article{dpo,
  author       = {Rafael Rafailov and
                  Archit Sharma and
                  Eric Mitchell and
                  Stefano Ermon and
                  Christopher D. Manning and
                  Chelsea Finn},
  title        = {Direct Preference Optimization: Your Language Model is Secretly a
                  Reward Model},
  journal      = {CoRR},
  volume       = {abs/2305.18290},
  year         = {2023},
  eprinttype    = {arXiv},
  eprint       = {2305.18290},
}

@article{hpl1,
  author       = {OpenAI},
  title        = {{GPT-4} Technical Report},
  journal      = {CoRR},
  volume       = {abs/2303.08774},
  year         = {2023},
  eprinttype    = {arXiv},
  eprint       = {2303.08774},
}

@article{entropy-metric,
  title={Detecting hallucinations in large language models using semantic entropy},
  author={Farquhar, Sebastian and Kossen, Jannik and Kuhn, Lorenz and Gal, Yarin},
  journal={Nature},
  volume={630},
  number={8017},
  pages={625--630},
  year={2024},
  publisher={Nature Publishing Group UK London}
}

@article{deepseekr1,
  author       = {DeepSeek{-}AI and
                  Daya Guo and
                  Dejian Yang and
                  et al.},
  title        = {DeepSeek-R1: Incentivizing Reasoning Capability in LLMs via Reinforcement
                  Learning},
  journal      = {CoRR},
  volume       = {abs/2501.12948},
  year         = {2025},
  eprinttype    = {arXiv},
  eprint       = {2501.12948},
}

@article{echolnk-r1,
  author       = {Zhenghao Xing and
                  Xiaowei Hu and
                  Chi{-}Wing Fu and
                  Wenhai Wang and
                  Jifeng Dai and
                  Pheng{-}Ann Heng},
  title        = {EchoInk-R1: Exploring Audio-Visual Reasoning in Multimodal LLMs via
                  Reinforcement Learning},
  journal      = {CoRR},
  volume       = {abs/2505.04623},
  year         = {2025},
  eprinttype    = {arXiv},
  eprint       = {2505.04623},
}

@inproceedings{rhlf2,
  author       = {James MacGlashan and
                  Mark K. Ho and
                  Robert Tyler Loftin and
                  Bei Peng and
                  Guan Wang and
                  David L. Roberts and
                  Matthew E. Taylor and
                  Michael L. Littman},
  editor       = {Doina Precup and
                  Yee Whye Teh},
  title        = {Interactive Learning from Policy-Dependent Human Feedback},
  booktitle    = {Proceedings of the 34th International Conference on Machine Learning,
                  {ICML}},
  pages        = {2285--2294},
  year         = {2017},
}

@article{ppo,
  author       = {John Schulman and
                  Filip Wolski and
                  Prafulla Dhariwal and
                  Alec Radford and
                  Oleg Klimov},
  title        = {Proximal Policy Optimization Algorithms},
  journal      = {CoRR},
  volume       = {abs/1707.06347},
  year         = {2017},
  eprinttype    = {arXiv},
  eprint       = {1707.06347},
}

@article{grpo,
  author       = {Zhihong Shao and
                  Peiyi Wang and
                  Qihao Zhu and
                  Runxin Xu and
                  Junxiao Song and
                  Mingchuan Zhang and
                  Y. K. Li and
                  Y. Wu and
                  Daya Guo},
  title        = {DeepSeekMath: Pushing the Limits of Mathematical Reasoning in Open
                  Language Models},
  journal      = {CoRR},
  volume       = {abs/2402.03300},
  year         = {2024},
  eprinttype    = {arXiv},
  eprint       = {2402.03300},
}

@inproceedings{Liu2025SuperRLRL,
  title={SuperRL: Reinforcement Learning with Supervision to Boost Language Model Reasoning},
  author={Yihao Liu and Shuocheng Li and Lang Cao and Yuhang Xie and Mengyu Zhou and Haoyu Dong and Xiaojun Ma and Shi Han and Dongmei Zhang},
  year={2025},
  url={https://api.semanticscholar.org/CorpusID:279075961}
}

@article{llama2,
  author       = {Hugo Touvron and
                  Louis Martin and
                  Kevin Stone and
                  Peter Albert and
                  Amjad Almahairi and
                  Yasmine Babaei and
                  Nikolay Bashlykov and
                  Soumya Batra and
                  Prajjwal Bhargava and
                  Shruti Bhosale and
                  Dan Bikel and
                  Lukas Blecher and
                  Cristian Canton{-}Ferrer and
                  Moya Chen and
                  Guillem Cucurull and
                  David Esiobu and
                  Jude Fernandes and
                  Jeremy Fu and
                  Wenyin Fu and
                  Brian Fuller and
                  Cynthia Gao and
                  Vedanuj Goswami and
                  Naman Goyal and
                  Anthony Hartshorn and
                  Saghar Hosseini and
                  Rui Hou and
                  Hakan Inan and
                  Marcin Kardas and
                  Viktor Kerkez and
                  Madian Khabsa and
                  Isabel Kloumann and
                  Artem Korenev and
                  Punit Singh Koura and
                  Marie{-}Anne Lachaux and
                  Thibaut Lavril and
                  Jenya Lee and
                  Diana Liskovich and
                  Yinghai Lu and
                  Yuning Mao and
                  Xavier Martinet and
                  Todor Mihaylov and
                  Pushkar Mishra and
                  Igor Molybog and
                  Yixin Nie and
                  Andrew Poulton and
                  Jeremy Reizenstein and
                  Rashi Rungta and
                  Kalyan Saladi and
                  Alan Schelten and
                  Ruan Silva and
                  Eric Michael Smith and
                  Ranjan Subramanian and
                  Xiaoqing Ellen Tan and
                  Binh Tang and
                  Ross Taylor and
                  Adina Williams and
                  Jian Xiang Kuan and
                  Puxin Xu and
                  Zheng Yan and
                  Iliyan Zarov and
                  Yuchen Zhang and
                  Angela Fan and
                  Melanie Kambadur and
                  Sharan Narang and
                  Aur{\'{e}}lien Rodriguez and
                  Robert Stojnic and
                  Sergey Edunov and
                  Thomas Scialom},
  title        = {Llama 2: Open Foundation and Fine-Tuned Chat Models},
  journal      = {CoRR},
  volume       = {abs/2307.09288},
  year         = {2023},
  eprinttype    = {arXiv},
  eprint       = {2307.09288},
}

@article{zou2024look,
  title={Look twice before you answer: Memory-space visual retracing for hallucination mitigation in multimodal large language models},
  author={Zou, Xin and Wang, Yizhou and Yan, Yibo and Lyu, Yuanhuiyi and Zheng, Kening and Huang, Sirui and Chen, Junkai and Jiang, Peijie and Liu, Jia and Tang, Chang and others},
  journal={arXiv preprint arXiv:2410.03577},
  year={2024}
}

@inproceedings{LAVISH,
  author       = {Yan{-}Bo Lin and
                  Yi{-}Lin Sung and
                  Jie Lei and
                  Mohit Bansal and
                  Gedas Bertasius},
  title        = {Vision Transformers are Parameter-Efficient Audio-Visual Learners},
  booktitle    = {{IEEE/CVF} Conference on Computer Vision and Pattern Recognition,
                  {CVPR}},
  pages        = {2299--2309},
  publisher    = {{IEEE}},
  year         = {2023},
}

@inproceedings{clip,
  author       = {Alec Radford and
                  Jong Wook Kim and
                  Chris Hallacy and
                  Aditya Ramesh and
                  Gabriel Goh and
                  Sandhini Agarwal and
                  Girish Sastry and
                  Amanda Askell and
                  Pamela Mishkin and
                  Jack Clark and
                  Gretchen Krueger and
                  Ilya Sutskever},
  editor       = {Marina Meila and
                  Tong Zhang},
  title        = {Learning Transferable Visual Models From Natural Language Supervision},
  booktitle    = {Proceedings of the International Conference on Machine Learning,
                  {ICML}},
  volume       = {139},
  pages        = {8748--8763},
  publisher    = {{PMLR}},
  year         = {2021},
}

@inproceedings{pstp-net,
  author       = {Guangyao Li and
                  Wenxuan Hou and
                  Di Hu},
  editor       = {Abdulmotaleb El{-}Saddik and
                  Tao Mei and
                  Rita Cucchiara and
                  Marco Bertini and
                  Diana Patricia Tobon Vallejo and
                  Pradeep K. Atrey and
                  M. Shamim Hossain},
  title        = {Progressive Spatio-temporal Perception for Audio-Visual Question Answering},
  booktitle    = ACMMM,
  pages        = {7808--7816},
  publisher    = {{ACM}},
  year         = {2023},
}

@article{CAD,
  author       = {Asmar Nadeem and
                  Adrian Hilton and
                  Robert Dawes and
                  Graham Thomas and
                  Armin Mustafa},
  title        = {{CAD} - Contextual Multi-modal Alignment for Dynamic {AVQA}},
  journal      = {CoRR},
  volume       = {abs/2310.16754},
  year         = {2023},
  eprinttype    = {arXiv},
  eprint       = {2310.16754},
}

@inproceedings{pope,
  author       = {Yifan Li and
                  Yifan Du and
                  Kun Zhou and
                  Jinpeng Wang and
                  Wayne Xin Zhao and
                  Ji{-}Rong Wen},
  editor       = {Houda Bouamor and
                  Juan Pino and
                  Kalika Bali},
  title        = {Evaluating Object Hallucination in Large Vision-Language Models},
  booktitle    = EMNLP,
  pages        = {292--305},
  publisher    = {Association for Computational Linguistics},
  year         = {2023},
}

@inproceedings{ye2024CAT,
  author       = {Qilang Ye and
                  Zitong Yu and
                  Rui Shao and
                  Xinyu Xie and
                  Philip Torr and
                  Xiaochun Cao},
  title        = {{CAT:} Enhancing Multimodal Large Language Model to Answer Questions
                  in Dynamic Audio-Visual Scenarios},
  booktitle    = {Computer Vision - {ECCV} 2024 - 18th European Conference, Milan, Italy,
                  September 29-October 4, 2024, Proceedings, Part {X}},
  volume       = {15068},
  pages        = {146--164},
  year         = {2024},
}

@article{qwenomni,
  author       = {Jin Xu and
                  Zhifang Guo and
                  Jinzheng He and
                  et al.},
  title        = {Qwen2.5-Omni Technical Report},
  journal      = {CoRR},
  year         = {2025},
  eprinttype    = {arXiv},
  eprint       = {2503.20215},
}

@article{gemini,
  author       = {Machel Reid and
                  Nikolay Savinov and
                  Denis Teplyashin and
                  et al.},
  title        = {Gemini 1.5: Unlocking multimodal understanding across millions of
                  tokens of context},
  journal      = {CoRR},
  volume       = {abs/2403.05530},
  year         = {2024},
}

@inproceedings{anygpt,
  author       = {Jun Zhan and
                  Junqi Dai and
                  Jiasheng Ye and
                  Yunhua Zhou and
                  Dong Zhang and
                  Zhigeng Liu and
                  Xin Zhang and
                  Ruibin Yuan and
                  Ge Zhang and
                  Linyang Li and
                  Hang Yan and
                  Jie Fu and
                  Tao Gui and
                  Tianxiang Sun and
                  Yu{-}Gang Jiang and
                  Xipeng Qiu},
  editor       = {Lun{-}Wei Ku and
                  Andre Martins and
                  Vivek Srikumar},
  title        = {AnyGPT: Unified Multimodal {LLM} with Discrete Sequence Modeling},
  booktitle    = {Proceedings of the 62nd Annual Meeting of the Association for Computational
                  Linguistics (Volume 1: Long Papers), {ACL} 2024, Bangkok, Thailand,
                  August 11-16, 2024},
  pages        = {9637--9662},
  publisher    = {Association for Computational Linguistics},
  year         = {2024},
}

@article{lin2025reliable,
  title={Reliable and Balanced Transfer Learning for Generalized Multimodal Face Anti-Spoofing},
  author={Lin, Xun and Liu, Ajian and Yu, Zitong and Cai, Rizhao and Wang, Shuai and Yu, Yi and Wan, Jun and Lei, Zhen and Cao, Xiaochun and Kot, Alex},
  journal={IEEE Transactions on Pattern Analysis and Machine Intelligence},
  year={2025},
  publisher={IEEE}
}

@article{ye2024pose,
  title={Pose-Promote: Progressive Visual Perception for Activities of Daily Living},
  author={Ye, Qilang and Yu, Zitong},
  journal={IEEE Signal Processing Letters},
  year={2024},
  publisher={IEEE}
}

@article{ye2024iet,
  author       = {Hai Nan and
                  Qilang Ye and
                  Zitong Yu and
                  Kang An},
  title        = {3sG: Three-stage guidance for indoor human action recognition},
  journal      = {{IET} Image Process.},
  volume       = {18},
  number       = {8},
  pages        = {2000--2010},
  year         = {2024},
  url          = {https://doi.org/10.1049/ipr2.13078},
  doi          = {10.1049/IPR2.13078},
}

@article{ye2025cat+,
  title={CAT+: Investigating and Enhancing Audio-visual Understanding in Large Language Models},
  author={Ye, Qilang and Yu, Zitong and Shao, Rui and Cui, Yawen and Kang, Xiangui and Liu, Xin and Torr, Philip and Cao, Xiaochun},
  journal={IEEE Transactions on Pattern Analysis and Machine Intelligence},
  year={2025},
  publisher={IEEE}
}

@article{omni-r1,
  author       = {Hao Zhong and
                  Muzhi Zhu and
                  Zongze Du and
                  Zheng Huang and
                  Canyu Zhao and
                  Mingyu Liu and
                  Wen Wang and
                  Hao Chen and
                  Chunhua Shen},
  title        = {Omni-R1: Reinforcement Learning for Omnimodal Reasoning via Two-System
                  Collaboration},
  journal      = {CoRR},
  volume       = {abs/2505.20256},
  year         = {2025},
  eprinttype    = {arXiv},
  eprint       = {2505.20256},
}

@inproceedings{avhbench,
  author       = {Kim Sung{-}Bin and
                  Oh Hyun{-}Bin and
                  JungMok Lee and
                  Arda Senocak and
                  Joon Son Chung and
                  Tae{-}Hyun Oh},
  title        = {AVHBench: {A} Cross-Modal Hallucination Benchmark for Audio-Visual
                  Large Language Models},
  booktitle    = {The Thirteenth International Conference on Learning Representations,
                  {ICLR} 2025, Singapore, April 24-28, 2025},
  year         = {2025},
}

\end{document}